\pdfoutput=1

\documentclass[11pt]{article}

\usepackage{acl}
\usepackage{booktabs}
\usepackage{times}
\usepackage{latexsym}
\usepackage{enumitem}
\usepackage[T1]{fontenc}

\usepackage[utf8]{inputenc}

\usepackage{microtype}

\usepackage{inconsolata}

\usepackage{graphicx}
\usepackage{subcaption}
\title{Striking a Balance between Classical and Deep Learning Approaches in Natural Language Processing Pedagogy}

\author{Aditya Joshi$^1$, Jake Renzella$^1$, Pushpak Bhattacharyya$^2$, Saurav Jha$^1$, Xiangyu Zhang$^1$ \\
  $^1$ University of New South Wales, Sydney, Australia \\
  $^2$ Indian Institute of Technology Bombay, Mumbai, India \\
  \texttt{$\{$aditya.joshi,jake.renzella,saurav.jha,xiangyu.zhang2$\}$@unsw.edu.au} \\
  \texttt{pb@cse.iitb.ac.in} \\}

\begin{document}
\maketitle
\begin{abstract}
While deep learning approaches represent the state-of-the-art of natural language processing (NLP) today, classical algorithms and approaches still find a place in NLP textbooks and courses of recent years. This paper discusses the perspectives of conveners of two introductory NLP courses taught in Australia and India, and examines how classical and deep learning approaches can be balanced within the lecture plan and assessments of the courses. We also draw parallels with the objects-first and objects-later debate in CS1 education. We observe that teaching classical approaches adds value to student learning by building an intuitive understanding of NLP problems, potential solutions, and even deep learning models themselves. Despite classical approaches not being state-of-the-art, the paper makes a case for their inclusion in NLP courses today.
\end{abstract}
\section{Introduction}
Transformer-based models are the state-of-the-art in natural language processing (NLP). They represent a new era of deep learning approaches\footnote{The phrase `deep learning approaches' is expected to refer to the broad spectrum including but not limited to dense word representations, Transformer and Transformer-based models used in NLP.} for NLP. From early work in word representations to recent approaches in instruction tuning, deep learning approaches have significantly transformed NLP and many other areas of artificial intelligence. Universities around the world have incorporated deep learning approaches in their introductory NLP courses, arguably with varying speed, and to different extents, as evidenced to an extent, in Table 2). This puts under question the relevance of classical approaches, \textit{i.e.}, those pre-dating deep learning approaches, in NLP curricula. This reflection from academics/faculty members may also be based on questions from students taking the course who wonder why they should learn about classical approaches, as seen in pre-course feedback received by the authors of this paper.

Therefore, we analyse the role of classical approaches in the context of modern university-based courses in NLP. We focus on introductory NLP courses offered to students of computer science \& engineering or equivalent degrees, and investigate the question:

``\textit{How can an introductory NLP course balance between the content covering classical and deep learning approaches?}"

We address the question in two parts: (a) how others do it, and (b) how we do it. With respect to (a), we describe summaries of NLP textbooks and publicly available courses in the context of the question. As for (b), we draw insights from our experience teaching NLP courses at two large, research-intensive universities in Australia and India. We introduce and discuss our motivations, considerations and decisions in the lectures, tutorials and projects of NLP courses. Authors of this paper represent two \textit{personas} of NLP educators: (a) an early-career educator who introduced a new NLP course in the beginning of 2024; and (b) a seasoned educator with two-decade experience of teaching and research experience. The novelty of this paper is as follows:
\begin{enumerate}
    \item There have been papers in the past on specific aspects (tutorials, lecture content, etc.) of individual NLP courses~\cite{plank-2021-back, foster-wagner-2021-naive, gaddy-etal-2021-interactive}. This paper is novel in its comparison of two NLP courses taught by educators with significantly different experience in NLP research and pedagogy.
    \item We also draw parallels to the age-long objects-first or -later debate in computer science education, to highlight that the classical approaches in NLP courses can benefit from lessons in computer science education.
    \item We hope that the paper serves as a useful resource for NLP educators to examine the role of classical methods in their curricula now and in the future.
\end{enumerate}

The rest of the paper is described as follows. We first introduce the context in terms of the two courses being compared in Section~\ref{sec:context}. Following that, we discuss an analogous objects first- and objects-later- debate in computer science education in Section~\ref{sec:analogies}. We describe how NLP textbooks and NLP courses (as per publicly available course outlines) cover classical approaches in Section~\ref{sec:books}. We then proceed to compare the lecture plan, and coding assessments involved in the two courses in Sections~\ref{sec:lectureplan} and ~\ref{sec:coding} respectively. Section~\ref{sec:makingthecase} puts it all together to make our case for the relevance of classical approaches in NLP courses. Finally, Section~\ref{sec:conclusion} concludes the paper.
\section{Context}
\label{sec:context}
The two NLP courses we compare are run at two large universities in Australia and India, both titled `Natural Language Processing'. We refer to these as \textbf{NLP-UNSW} and \textbf{NLP-IITB}\footnote{UNSW: University of New South Wales, Sydney, Australia; IITB: Indian Institute of Technology Bombay, Mumbai, India.}. NLP-UNSW is the first offering of the course with 60 enrolled students, while NLP-IITB is the ninteenth offering of the course with 150 enrolled students. While NLP-UNSW is the only NLP-focussed course at UNSW at the time of writing this paper, NLP-IITB is accompanied with a `Deep learning for NLP' course which often follows NLP-IITB. Both courses were offered as an undergraduate/postgraduate elective as a part of the computer science and engineering programme, and were introductory courses to NLP. The cohorts consisted of undergraduate (nearly 50\%) and postgraduate students. Postgraduate students included those enrolled in Masters by Coursework or Masters by Research programmes, and early PhD students.

\begin{figure}
    \centering
    \includegraphics[width=0.45\textwidth]{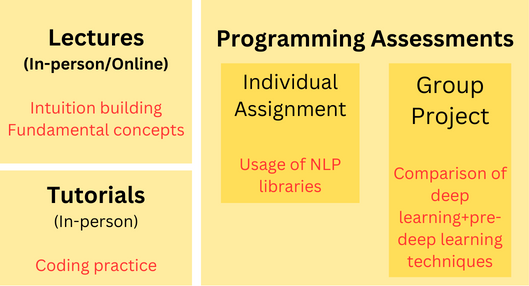}
    \caption{Course Structure for NLP-UNSW.}
    \label{fig:coursecomponentsa}
\end{figure}

\begin{figure}
    \centering
    \includegraphics[width=0.45\textwidth]{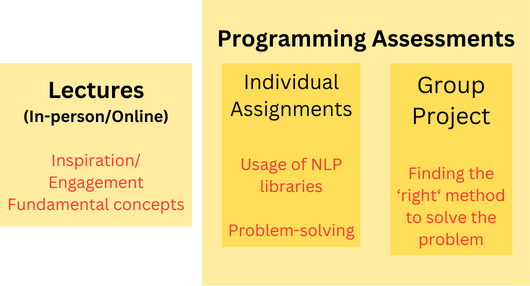}
    \caption{Course Structure for NLP-IITB.}
    \label{fig:coursecomponentsb}
\end{figure}

The course components of NLP-UNSW and NLP-IITB, along with the course conveners' reasons behind covering classical approaches, are shown in Figures~\ref{fig:coursecomponentsa} and ~\ref{fig:coursecomponentsb} respectively. Generally speaking, an exposition to classical approaches allowed the students to solve problems and understand deep learning approaches vis-\'a-vis their predecessors. NLP-UNSW insisted that the group project compare deep learning methods with their predecessors. NLP-IITB allowed the students to choose an appropriate method. In contrast, the individual assignment in NLP-UNSW only involved using black-box NLP libraries while NLP-IITB used multiple assignments covering both statistical as well as deep learning models.

\section{Parallels to Computer Science Education}
\label{sec:analogies}
As NLP educators grapple with Classical-first, -later, or perhaps -interleaved approaches to curricula, it is helpful to draw upon empirical research and lessons learned by CS1 educators (introductory programming) who have grappled with a similar dilemma: the age-long objects-first or -later debate. For decades, introductory programming students learned imperative-style programming in languages such as C, Pascal, and Fortran. In the early 2000s, educators started incorporated object-oriented programming concepts into their curriculum as object-oriented programming took hold in industry \cite{Cooper2003}. Where the objects-first community may begin immediately with concepts of composition, inheritance, and abstraction, the objects-later community instead focus on simpler, foundational programs and concepts such as sequence and control flow.

The primary criticism of objects-first approaches surrounds the added complexity necessary to discuss object-oriented concepts \cite{Proulx2002}. Objects-first must tackle objects, classes, encapsulation, and access modifiers to discuss even the most basic object-oriented concepts. Of course, objects-first educators will retort that this extra effort is worth it, with students reporting a solid sense of program design and contextualisation of object oriented concepts \cite{Cooper2003}.

These criticisms of object-first approaches are not unfounded. We can apply cognitive load theory, the theory of human cognition and how learning occurs, to inform instructional design in computing and NLP. The theory, backed by many empirical studies, finds that limited amounts of secondary information (such as computing tasks) can be processed at any given time \cite{Sweller2011}. Put simply, human cognitive architecture lends itself to learning best when concepts are minimally introduced. This reduces the cognitive overload that can hinder the assimilation of new information.

Cognitive load theory has been applied in introductory programming contexts, finding that cognitive load measures and tools can be applied to computing education to inform curriculum and instructional design \cite{Morrison2014}.

While cognitive load theory may, at first glance, support an objects-later approach due to the reduction of concepts required, approaches to reduce cognitive load when delivering objects-first curriculum may be preferable if it aligns with its goals.

Empirical studies of learning outcomes between objects-first and -later seek to focus on measurable evaluations of student learning outcomes. \citet{Ehlert2009} found no differences in learning gain between objects-first and objects-later courses; however, they did find that students in objects-later courses reported lower perceived difficulty and higher comfort levels. \citep{Tew2005} compared objects-first/later at the same institution and found that students performed comparatively by the end of their second term. A more recent, large-scale systematic literature review on Introductory Programming seems to support this nuanced take on programming paradigms \cite{LuxtonReilly:2018ko}. The authors find that there is still active research on programming paradigms, and surmise with: ``after full consideration of all the papers it is by no means clear that any paradigm is superior to any other for the teaching of introductory programming". We may glean from this comparison to CS1 education design that many dimensions are involved in curriculum design. Empirical studies have found that the goals of the program and the institution itself, perspectives of educators, and quality of instruction and instructional materials are more critical than an objects-first or -later design.

Unlike CS1 education, however, NLP may often be available as a single elective in a computer science program. Therefore, NLP educators are forced to balance providing practical, deep learning skills to students while also providing solid foundations in classical NLP in a shorter time-frame, perhaps supporting an interleaved approach to NLP curricula.
\begin{table*}[]
\begin{tabular}{p{4cm}p{0.5cm}p{6.5cm}p{3.5cm}}
\toprule
Authors; Publisher & Year & Chapter Layout & Deep learning focus    \\ \midrule

 Speech and Language Processing; Dan Jurafsky, James H Martin; - & 2024 & Fundamental Algorithms (Statistical Models, Neural Models, RNNs, LSTMs.) followed by NLP applications and linguistic tasks. & Techniques first. classical: Yes.  \\ 
A Course in Natural Language Processing; Yannis Haralambous; Springer & 2024 & Layers of NLP: Phoneme, Grapheme, Morpheme. Last chapter: Going Neural.  & Neural Models (RNNs, Transformers) \\ %
Natural Language Processing; Pushpak Bhattacharyya, Aditya Joshi; Wiley & 2023 & Fundamental techniques first. Then neural. Significant focus on NLP   problems.    & Interleave classical and neural both. Three generations. \\ %
Real-World Natural Language Processing; Masato Hagiwara; Manning      & 2022 & Neural-focussed. Embeddings, SentenceClasification, Seq2Seq, Transformer, etc.  & Neural-focussed.     \\ %
Deep Learning for Natural Language   Processing; Stephan Raaijmakers; Manning Publications & 2022 & Text embeddings, sequential NLP, attention, multitask learning, last chapter: Transformers and using Transformers. Lots of code examples            & Neural focussed.                                                    \\%
Practical Natural Language Processing; Sowmya   Vajjala, Bodhisattwa Majumder, Anuj Gupta, Harshit Surana; O'Reilly             & 2020 & NLP Primers: Quick introduction to primary NLP concepts. Very task driven:   Classification, IE, Chatbots, Applications to domains                            & Cover deep learning Early. Focus on applications and tasks          \\ %
Introduction to Natural Language   Processing; Jacob Eisenstein; MIT Press            & 2019 & Learning: Classification (Tasks and approaches); Sequences (Tasks and approaches); Semantics. Word embeddings, parsing, reference resoluton                & Pre-deep learning first. Then neural.                               \\ %
\bottomrule                                                                                                                       
\end{tabular}
\caption{Deep learning focus in some NLP textbooks listed in the reverse-chronological order.\label{tab:textbooks}}
\end{table*}
\section{Observations from NLP textbooks and courses}
\label{sec:books}
Table~\ref{tab:textbooks} presents a summary of some NLP textbooks, in terms of layouts and their deep learning focus. The books were identified using a search on the UNSW library. The focus was identified using the content plan of the book, along with a surface analysis of the chapters. We observe that nearly all textbooks cover classical approaches, primarily in terms of statistical models. We describe two textbooks in particular. The textbook by Jurafsky-Martin~\footnote{\url{https://web.stanford.edu/~jurafsky/slp3/}} divides its chapters into fundamental algorithms and NLP applications. The first five chapters introduce regular expressions, one-hot vectors and statistical algorithms like support vector machines and logistic regression. In the subsequent chapters, the book covers word embeddings, then begins to combine LSTM with CRFs in the context of POS tagging. Following that, the book covers Transformer, fine-tuning and prompting/prompt learning Transformer. \citet{ournlpbook} use a different approach. They cover fundamental algorithms for representation learning: computational grammar, probabilistic language modeling, and word2vec/LSTM-based representations. The book visualizes NLP as three generations: rule-based, statistical and neural. Following that, the book introduces Transformer. The subsequent chapters alternate between the approaches in the three generations.

Table~\ref{tab:courses} illustrates how some NLP courses cover classical topics. We only use examples of courses whose course outlines are publicly available on the internet. Similarly, this is not a complete list either. In universities where there is a deep learning-focused course for NLP, topics such as syntactic parsing or statistical classification are covered in substantial detail. Even for courses that cover little classical content, n-gram language modeling is a topic that is included. The table also shows `when is Transformer taught' which may be viewed as a turning point when deep learning-based approaches assume focus in the course. We view Transformer as a transformative technology in NLP, and do not claim that all of deep learning is Transformer.

\begin{table*}[]
\begin{tabular}{p{2cm}p{1cm}p{6cm}p{5cm}}
\toprule
\textbf{University}      & \textbf{Year} & \textbf{Examples of pre-deep-learning topics}                   & \textbf{When is Transformer taught?}                                       \\ \midrule
UMAss Amherst\footnote{\url{https://people.cs.umass.edu/~miyyer/cs685/}}        & 2023          & N-gram language modeling                                        & Week 2 of 11             
\\
CMU\footnote{\url{https://phontron.com/class/anlp2024/course_details/}}                  & 2024          & Representing words                                              & Week 3 of 16                                                                   \\
IIT Delhi\footnote{\url{https://www.cse.iitd.ac.in/~mausam/courses/col772/spring2024/}}          & 2022          & Finite state automata, statistical parsing                      & End of month 2 out of 3.5                                                      \\
University of Edinburgh\footnote{\url{http://www.drps.ed.ac.uk/23-24/dpt/cxinfr10078.htm}} & 2024          & Lexical, syntactic and semantic parsing, etc.                   & This only covers algorithmic fundamentals; separate course for neural   models \\
University of Washington\footnote{\url{https://safe-fernleaf-26d.notion.site/Winter-24-CSE-447-517-Natural-Language-Processing-4142333a001143d2be5ecff1a535c4ab}}& 2024          & N-gram language modeling                                        & Week 2 of 11                                                                   \\
NYU  \footnote{\url{https://nyu-cs2590.github.io/spring2023/calendar/}}                   & 2023          & Feature-based classification, N-gram language modeling          & Week 3 of 15                                                                   \\
\bottomrule
\end{tabular}
\caption{Examples of the coverage of deep learning in NLP courses.\label{tab:courses}}
\end{table*}


\section{Lecture Plan}
\label{sec:lectureplan}

\begin{figure*}
    \centering
    \includegraphics[width=0.7\textwidth]{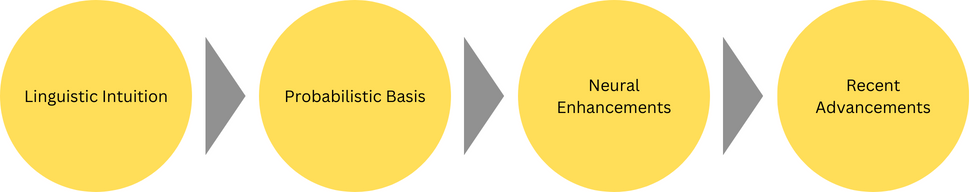}
    \caption{Lecture Plan for NLP-UNSW.}
    \label{fig:lecplana}
\end{figure*}

\textbf{NLP-UNSW}: In the NLP-UNSW course which is spread over 10 weeks, we adopt a hybrid methodology where we interleave deep learning approaches with statistical/rule-based approaches. The hybrid methodology is illustrated in Figure~\ref{fig:lecplana}. In the first week, we introduce NLP via black-boxes such as spacy, nltk, and HuggingFace pipelines. The students are introduced to NLP tasks with demonstrations of the three libraries. This is to help the students develop an understanding of the task. Similarly, if students aim to only `use' NLP in their projects (which may be sufficient for interdisciplinary projects), these libraries are sufficient. The focus here is also to help them understand the input and output of NLP systems, which we believe is the starting point to understanding NLP.

In week 2, we cover representations of text via one-hot vectors and probabilistic language modeling (statistical approach), and word representation learning as in word2vec and GloVe (deep learning approach). This not only allows students to appreciate the value addition of deep learning approaches but also identify situations in which non-deep learning approaches may be sufficient. In week 3, we focus on Transformer architecture: exposing the students to the architectural details, pseudocode, and code implementations. This is followed by Transformer-based models (encoder, decoder, and encoder-decoder models) in week 4. With this background in neural NLP models, we switch gears and focus on one NLP task every week. Every task is covered using the following steps: the linguistic task and associated challenges, classical approaches, deep learning approaches, and recent advances in the area. The first two allow the students to gain a foundational understanding of the problem, followed by the state-of-the-art in deep learning. The recent advances are catered to students who might be interested in future research in NLP without going into too much depth - to ensure that the content is accessible. The NLP tasks we cover are: sentiment analysis (representing sequence classification), named entity recognition (representing token classification), machine translation (representing sequence-to-sequence tasks), summarization (modified sequence-to-sequence tasks) and bias mitigation. 

In NLP-UNSW, we experience a recurring challenge when discussing NLP tasks every week. Foundation models (both encoder-only and decoder-only models) are versatile in their utility for different tasks. Therefore, we structure each neural section in three steps: (a) fine-tuning BERT, (b) one or more advanced methods relevant for the task. This serves two purposes. It allows us to teach different fine-tuning techniques. In addition, it also permits us to ground them in specific NLP tasks. We remember to remind students that the method is applicable in other NLP tasks.

\begin{figure}
    \centering
    \includegraphics[width=0.45\textwidth]{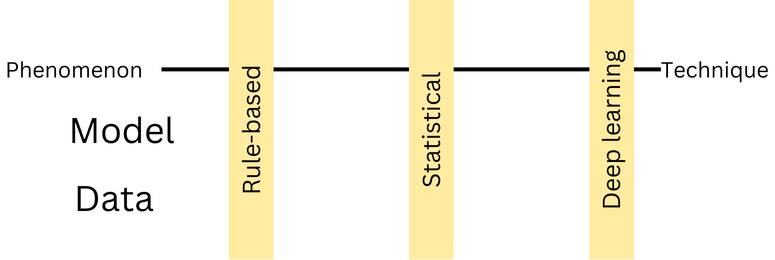}
    \caption{Lecture Plan for NLP-IITB.}
    \label{fig:lecplanb}
\end{figure}

\textbf{NLP-IITB}: NLP-IITB starts with a focus on sequence labeling via HMM-based POS tagging, followed by tree extraction via probabilistic parsing. In both cases, the mathematical details (driven by probability) go hand in hand with the algorithms (explained through the code). Then, machine translation is introduced, which makes way for an introduction to Transformer and large language models. The subsequent weeks discuss sentiment analysis as a specific NLP task and evaluation metrics in NLP. It must be noted that NLP-IITB is the introductory NLP course while there is a deep learning-specific NLP course at IITB which is often a follow-up to the introductory NLP-IITB. This is not the case for NLP-UNSW (being the only NLP-centric course). As a result, the course needs to balance between the two. This comparison shows how educators may make different choices, depending on the length of the teaching period. As a result, the lecture plan of NLP-IITB uses the spectrum shown in Figure~\ref{fig:lecplanb}. classical approaches allow focusing on the linguistic phenomenon which is captured using a model. The data provides the parameters of the model.

\textbf{Comparison}: In both NLP-UNSW and NLP-IITB, however, we observe that the foundational material serves as a good basis to introduce terminology (for example, types of summarization are abstractive and extractive summarization) and tag sets (as in the case of named entity recognition). The classical approaches build an intuitive understanding of building computational models for the specific NLP task. Appreciating the challenges in the classical approaches leads the students to the need for deep learning approaches. Through handwritten examples in the lecture, we highlight the relationships between the mathematics of classical approaches and Transformer-based approaches (for example, alignment model in statistical MT with cross-attention in Transformer).

\section{Coding Assessments}
\label{sec:coding}

Both NLP-UNSW and NLP-IITB have individual and group coding assessments. NLP-UNSW has one of each, while NLP-IITB has two individual and one group assessment(s).
\subsection{Course Assignment}
\textbf{NLP-UNSW}: The first coding assessment in NLP-UNSW is an individual coding assignment completed by students by the end of week 3. The coding assessment is run in a competition-like environment. The students are given a document defining the problem definition, and a template code that also contains stubs for testing. The students are required to add to the template. Closer to the submission, the students are given test cases to evaluate their code. The topic of the individual assignment needed deliberation. Until that point, we have only started discussing Transformer in the lectures. Therefore, we utilize the individual assignment to assess the classical skills of the students. In the case of this offering, the task was to extract skills from job ads, based on a skills ontology. The assignment was designed in two parts: the first part required the students to use the spacy matcher, while the second part required them to use embedding similarities. The former tests them for their ability to use a classical library while the latter requires them to use HuggingFace models.  As a result, the assessment covers the students' ability to `use' NLP models. Marking the assignment includes an automatic marking component with test cases and manual marking by the tutors. Being a rule-based system, the evaluation of test cases is found to return low precision. Therefore, rather than scoring on individual test cases, we mark the students on precision thresholds. 

\textbf{NLP-IITB}: NLP-IITB follows a similar strategy although there are multiple individual assignments. The course convener found it useful to give classical assignments on topics such as POS tagging, parsing and so on. An innovative assignment was also tried out, which met with good student feedback. Given a large dataset of names of cities, the task was to identify a preferred suffix in the name of a city. Students reported multiple deep learning/non-deep learning approaches.

\subsection{Course Project} 
Running in its first year, NLP-UNSW provides detailed guidelines for the course project. In contrast, NLP-IITB provides less guidelines but involves periodic consultation with the course team.

\textbf{NLP-UNSW}: The group project is a centerpiece of the NLP-UNSW course, allowing students to explore diverse topics within NLP. Teams of four to five students are encouraged to select a specific area of interest, ranging from NLP topics of active research such as prompt recovery in large language models to purely applied topics like developing a virtual learning assistant using Retrieval Augmented Generation \cite{lewis2020retrieval}. To guide students effectively, we provide a detailed scope guideline document. The document outlines the following key aspects alongside their respective credits (cr):

\begin{enumerate}[leftmargin=*]
    \item Problem definition: Must be an NLP problem ($5$ cr) with a text-based source/domain ($5$ cr);

    \item Dataset selection: Use an existing dataset ($10$ cr), create your own labelled dataset ($20$ cr) and use an existing lexicon ($10$ cr);

    \item Modelling: Implement a rule-based/statistical baseline ($50$ cr), use an existing pre-trained model ($5$ cr), fine-tune your own model ($50$ cr), and integrate a language model with external tools ($20$ cr);

    \item Evaluation: Quantitative ($10$ cr), qualitative ($5$ cr), command-line testing ($5$ cr) and demonstration ($10$ cr).
\end{enumerate}
Project teams are expected to cover a minimum of 100 credits. The aspects show that, while deep learning-based techniques, carry a significant focus, students are encouraged to experiment with simpler models.
\paragraph{Project evaluation.} The submission evaluation is structured around specific questions that capture the essence of a well-rounded NLP project pipeline. These include the evaluation of the project report -- for scope: architectural, methodological, and analytical details; group presentation -- for quality: architectural details, presentation format/style; codebase -- for code style: readability, scope, errors, and structure, and individual effort of group members. For the lattermost evaluation, each group is to submit an individual contribution file enlisting the technical contributions of each member.

\begin{figure}[htbp]
    \centering
    \begin{subfigure}[b]{0.23\textwidth}
        \centering
        \includegraphics[width=\textwidth]{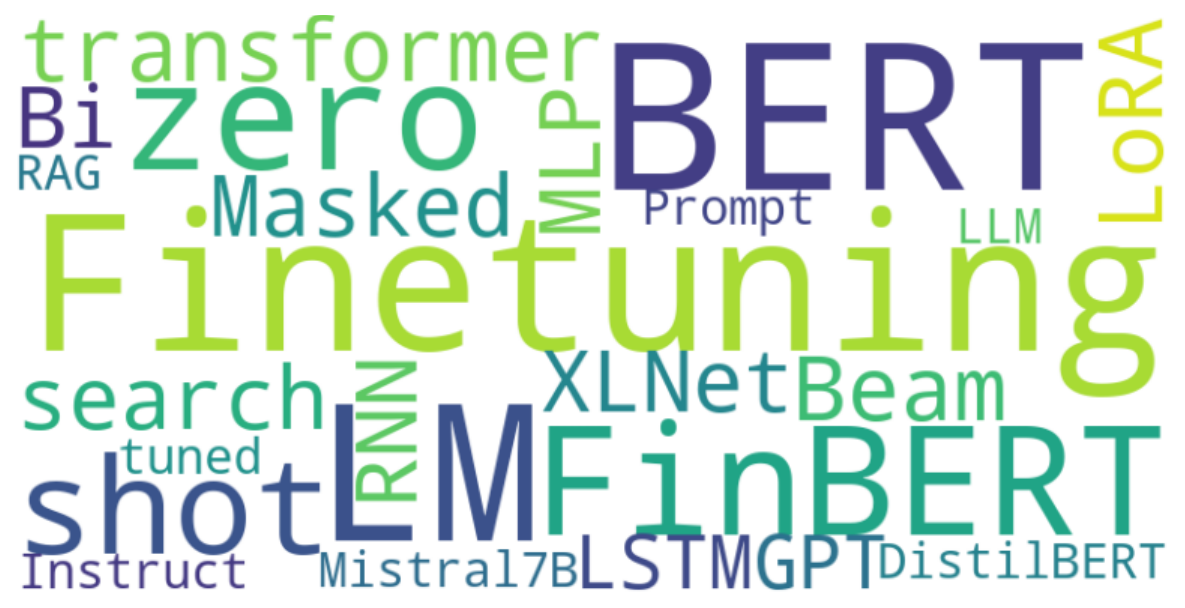}
        \caption{Deep-learning}
        \label{fig:subfig1}
    \end{subfigure}
    \hfill
    \begin{subfigure}[b]{0.23\textwidth}
        \centering
        \includegraphics[width=\textwidth]{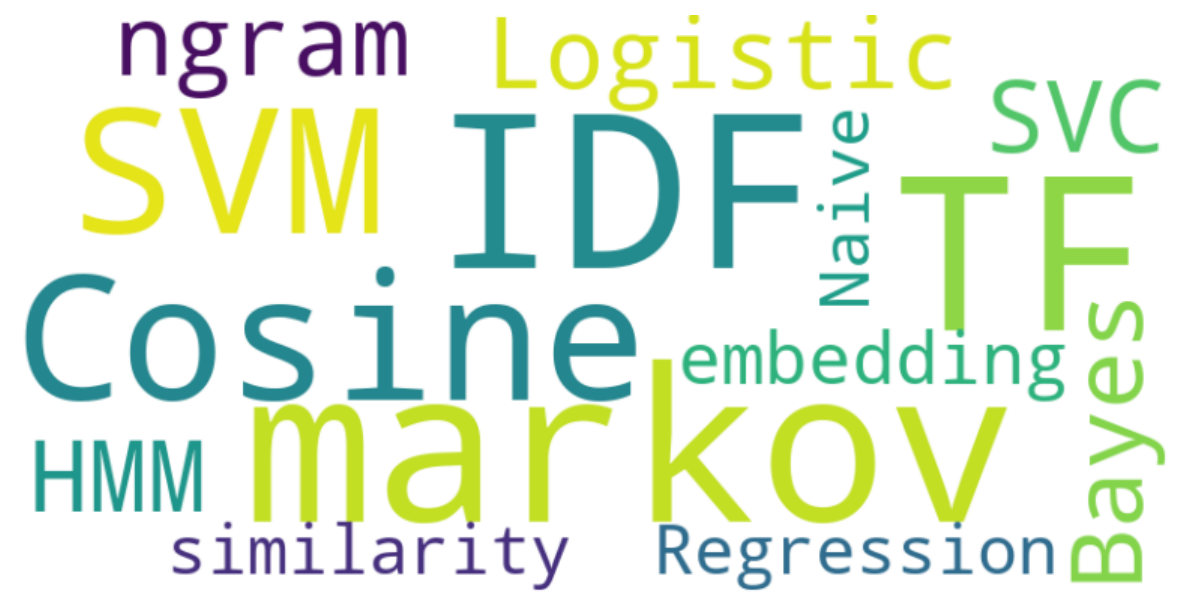}
        \caption{classical}
        \label{fig:subfig2}
    \end{subfigure}
    \caption{Word clouds showing the key deep learning and classical methods explored in the group projects.}
    \label{fig:bothfigures}
\end{figure}

\paragraph{Findings.} The diversity of modeling algorithms is particularly interesting across the submissions (see Fig. \ref{fig:bothfigures}). We note that classical techniques such as such as lexicon-based systems, traditional machine learning algorithms (e.g., SVM, Naive Bayes), and feature engineering-based techniques serve towards foundation building, interpretable analyses, and resource efficiency all while allowing the students to appreciate the advancements brought about by deep learning in NLP by benchmarking against them. Deep learning models (Fig. \ref{fig:subfig1}), on the other hand, help achieve complex pattern recognition, state-of-art performance, and end-to-end learning while helping the students understand the laws of scalability on large-scale datasets.

\textbf{NLP-IITB}: In NLP-IITB, we list a few topics for students to choose from, while being open to new ideas. Providing topics helps to streamline project ideas while giving student teams a headstart. The projects are evaluated using a combination of demonstration and presentation, including an in-person discussion with the course team. The focus is the selection of the `right' algorithm. The difference in the level of detail appears to reflect the experience of the two academics running the courses.
\subsection{Tutorials}

Tutorials are a weekly activity in NLP-UNSW (while not a part of NLP-IITB). In a typical tutorial, we first cover a focused review of the content covered in the lectures. We initially provide a brief overview of classical methods, as these approaches often greatly assist students in understanding the task. Among the classical topics, students show interest in learning about POS tagging and HMM-related algorithms. In the tutorial, we emphasize the mathematical derivation of HMM itself to improve the student's understanding. Subsequently, we extend the discussion by integrating the latest research related to the lecture topics. This part is focused on expanding on the lecture content to cover recent, trending papers that are not included in the syllabus. For instance, when discussing Parameter-Efficient Fine-Tuning (PEFT), we incorporate a paper-reading session of the paper by~\cite{he2021towards}, explaining its essence from a unified perspective and how it can be generally summarized. This allows us to meet the expectations of students who may be interested in advanced topics. Next, we utilize code demonstrations to illustrate the concepts presented in the lectures, facilitating a deeper understanding for the students. Following this, we explain and demonstrate the coursework requirements, such as homework assignments. Finally, we conclude by addressing student questions, which may pertain to projects or other assignments. Particularly in the tutorials, the students show less interest in the classical sections, focusing more on deep learning-based methods that are perceived as more employable. Students who actively participate and show interest in the derivation of various model principles are often also interested in research.

\section{Making the Case for Classical Approaches}
\label{sec:makingthecase}
Based on the considerations discussed so far, we are now able to make several arguments that may influence the inclusion of classical approaches in NLP courses. Designing a curriculum based on course expectations along these arguments may help determine the `balance' that is the focus of this paper.
\begin{figure}
    \centering
    \includegraphics[width=0.27\textwidth]{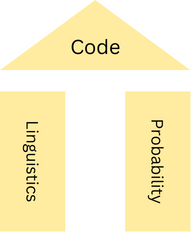}
    \caption{Intuition-building using classical approaches.}
    \label{fig:spectrum}
\end{figure}

\subsection{Intuition-building}

The conveners of NLP-UNSW and NLP-IITB find that rule-based and statistical approaches are great for intuition building and also appreciate that NLP is challenging. Classroom exercises where students describe rules for a particular NLP task enable them to see why a good rule-based system would be laborious due to the well-known high-precision-low-recall setup. Statistical approaches help to highlight the importance of probability in NLP. Probability is the cornerstone of NLP models, including modern models. Softmax being the center of Transformer-based models is an example of that. Starting with Bayes theorem, a teacher can effectively tease out dependencies between different variables, serving to be a great explanation to lead to attention and related concepts, especially during lectures. 

A pre-course survey done in NLP-IITB revealed that, out of 150 students, 80\% preferred that the course build intuition as opposed to providing information. Information refers to providing a suite of approaches and methods, while intuition refers to the motivation (linguistic or mathematical, etc.) underlying the methods. In fact, both courses combine the pillars illustrated in ~\ref{fig:spectrum}. Linguistics motivates the problem, probability and code run as common threads allowing a comparison between the two kinds of approaches. This motivation is highlighted by the instructor of NLP-IITB, by leveraging examples from his rich research experience. In contrast, the instructor of NLP-UNSW prefers to give examples from industry applications, given their professional experience in the tech industry.
\subsection{Student Motivation}
Several blog articles on NLP are available online. A clear differentiator in a university-based course is inspiring students using blended learning~\cite{deng2010exploring}. Towards this, classical approaches are effective in inspiring students to take up NLP projects. The conveners of NLP-UNSW and NLP-IITB observe that the challenges of rule-based systems are clear with some examples, during lectures. Covering classical approaches serves as a good motivator for students to see why deep learning approaches have been revolutionary to the field. This feeling of inspiration is often reported in NLP-IITB.
\subsection{Popular classical approaches}
Classical approaches have done well for linguistic tasks, particularly at the lower levels of the NLP hierarchy. Approaches like HMM and CRF work well for POS tagging and other token classification tasks. We observe that past courses also tend to cover classical methods as an introduction to deep learning methods. Comparison of popular classical approaches with deep learning approaches can aid learning during projects and assignments, as described in Section 6.
\subsection{Annotation}

Classical techniques, particularly in terms of annotation, are the benchmark for modern NLP models since tag sets used in classical approaches are still useful for their linguistic rigour. An example is the POS tagset. Nuances in Penn TreeBank about tags for ``JJ" versus ``JJR" (adjectives and comparative adjectives) help the students understand why they were designed in a certain way. The pre-course survey in NLP-IITB revealed that 67\% students preferred fundamental approaches versus state-of-the-art. The course covered a combination of the two.
\subsection{Cognitive load theory}

Classical techniques often involve explicit, well-defined rules and simpler models that can be more transparent and interpretable than their neural counterparts. This clarity can reduce extraneous cognitive load for learners by providing simpler examples of how inputs are transformed into outputs. For instance, a decision tree for language processing allows learners to see the exact paths through which decisions are made, thus aligning with the segmenting principle of cognitive load theory.

\section{Conclusion}
\label{sec:conclusion}
Classical approaches for NLP refer to those prior to the predominant use of neural networks and deep learning. This paper investigated the role of classical approaches in an introductory NLP course, by comparing perspectives from two courses: one which was offered for the first time and another which has been running for nineteen years while being continually revised. We discussed key considerations and reasons why classical approaches may be helpful, in terms of the past textbooks, lectures, and assessments. An understanding of classical NLP not only builds a strong foundation for the students but also enables them to look back at some of these methods to develop new techniques. We hope that the analyses presented in the paper will allow NLP educators and students alike to find the right balance between the classical and deep learning approaches.
\section*{Limitations}
The books and courses described in the paper are a subset curated only as an example. It does not represent a complete list. Similarly, the sections derive from insights teaching the two courses covered in the paper, and are not necessarily prescriptive.
\section*{Ethical Considerations}
The paper describes summary statistics of student surveys without identifying individual students or student cohorts. The assessment section provides only a high-level view of the assignments and lectures, and may not result in implications to academic integrity in the future versions of the courses.
\section*{Acknowledgment}
The authors thank Dipankar Srirag, UNSW Sydney, for his feedback on a near-camera-ready draft of the paper. 
\bibliography{acl_latex}
\bibliographystyle{acl_natbib}
\end{document}